\documentclass[final]{cvpr}

\usepackage{epsfig}
\usepackage{graphicx}
\usepackage{amsmath}
\usepackage{amssymb, eucal}
\usepackage{bbm,nicefrac}
\usepackage{makecell, verbatim, sidecap}
\usepackage[position=bottom]{subfig}

\usepackage[margin=6pt,font=small,labelfont=bf,labelsep=endash,tableposition=top]{caption}

\usepackage[pagebackref=true,breaklinks=true,colorlinks,bookmarks=false]{hyperref}

\usepackage{bm}
\newcommand{\myparagraph}[1]{{\vspace{.5em} \noindent \bf #1}}

\renewcommand{\myparagraph}[1]{{\vspace{.1em} \noindent \bf #1}}

\newcommand{\app}{\raise.17ex\hbox{$\scriptstyle\sim$}}
\newcommand{\tablestyle}[2]{\setlength{\tabcolsep}{#1}\renewcommand{\arraystretch}{#2}\centering\footnotesize}

\def\Ours{{DenseCL}\xspace}

\begin{document}

\title{Dense Contrastive Learning for 
Self-Supervised
Visual Pre-Training}

\author{
    Xinlong Wang$ ^1$, 
    ~ ~ ~ ~
    Rufeng Zhang$ ^2$,
    ~ ~ ~ ~
    Chunhua Shen$ ^1$\thanks{Corresponding author.},
    ~ ~ ~ ~
    Tao Kong$ ^3$,
    ~ ~ ~ ~
    Lei Li$ ^3$
    \\[0.2cm]
    $ ^1$The University of Adelaide, Australia
    ~ ~ 
    $ ^2$Tongji University, China
    ~ ~
    $ ^3$ByteDance AI Lab
}

\maketitle

\begin{abstract}

To date, most existing self-supervised learning methods are designed 
and optimized for image classification. These pre-trained models can be sub-optimal for dense prediction tasks due to the discrepancy between image-level prediction 
and pixel-level prediction.  To fill this gap, we aim to design
an effective, dense self-supervised learning method that directly works at the 
level of  pixels (or local features) by taking into account the correspondence between local features. 
We present dense contrastive learning (DenseCL), which implements self-supervised learning 
by optimizing a pairwise contrastive 
(dis)similarity
loss at the pixel level between two views of input images. 

Compared to the baseline method MoCo-v2, our method introduces negligible computation overhead (only $<$1\% slower), but demonstrates consistently superior performance when transferring to downstream dense prediction tasks including object detection, semantic segmentation and instance segmentation; and outperforms the state-of-the-art methods by a large margin.
Specifically, over the strong MoCo-v2 baseline, our method achieves significant improvements of 2.0\% AP on PASCAL VOC object detection, 1.1\% AP on COCO object detection, 0.9\% AP on COCO instance segmentation, 3.0\% mIoU on PASCAL VOC semantic segmentation and 1.8\% mIoU on Cityscapes semantic segmentation.

{   
    \def\UrlFont{\sf}
    \def\UrlFont{\rm\small\ttfamily}
Code and models are available at: \url{https://git.io/DenseCL} 
}

\end{abstract}

\begin{figure}[ht!]
    \centering
    \subfloat[Object Detection]{
        \includegraphics[width=0.22\textwidth]{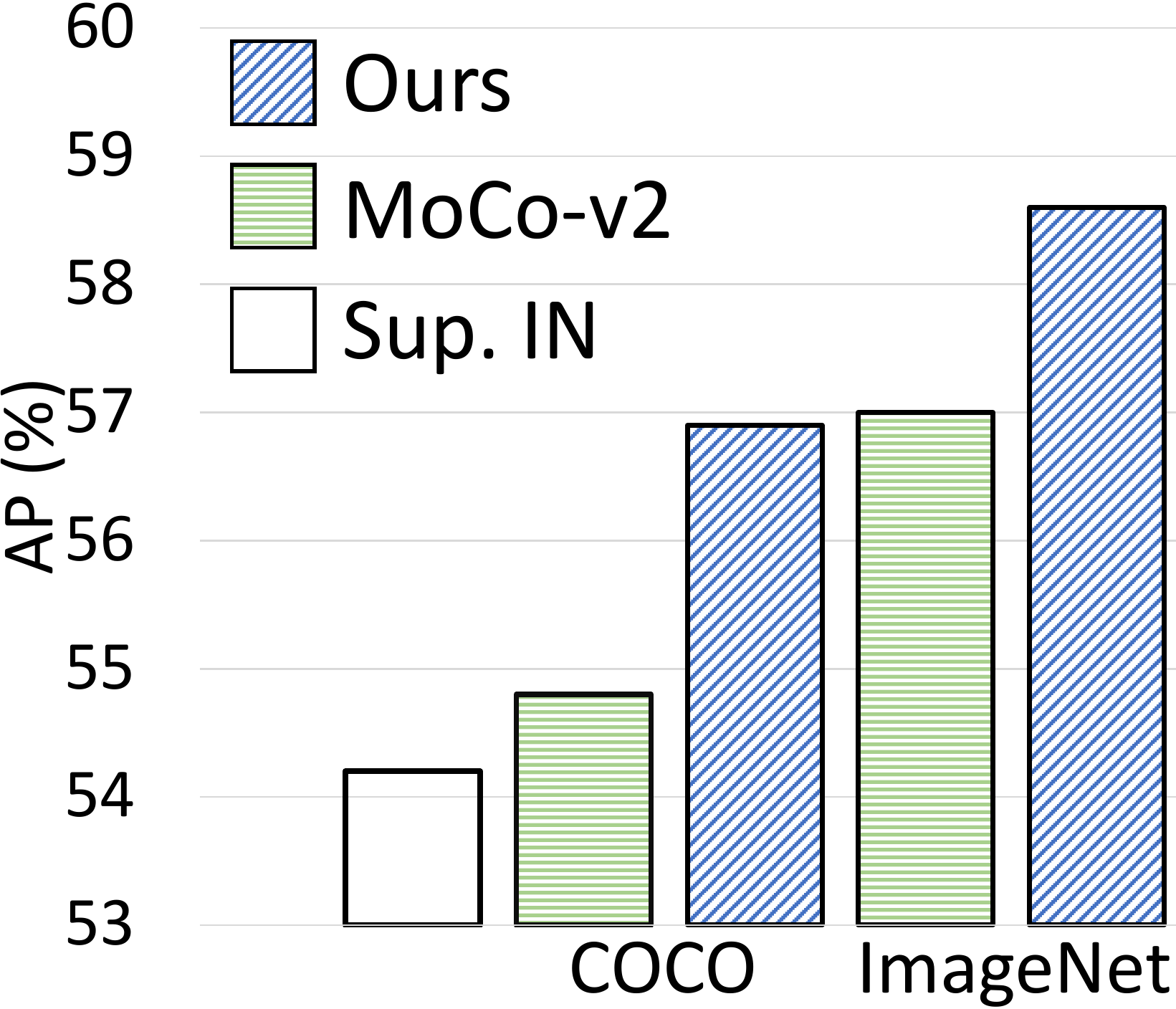}
        \label{fig:gains_det}
    }
    \subfloat[Semantic Segmentation]{
        \includegraphics[width=0.22\textwidth]{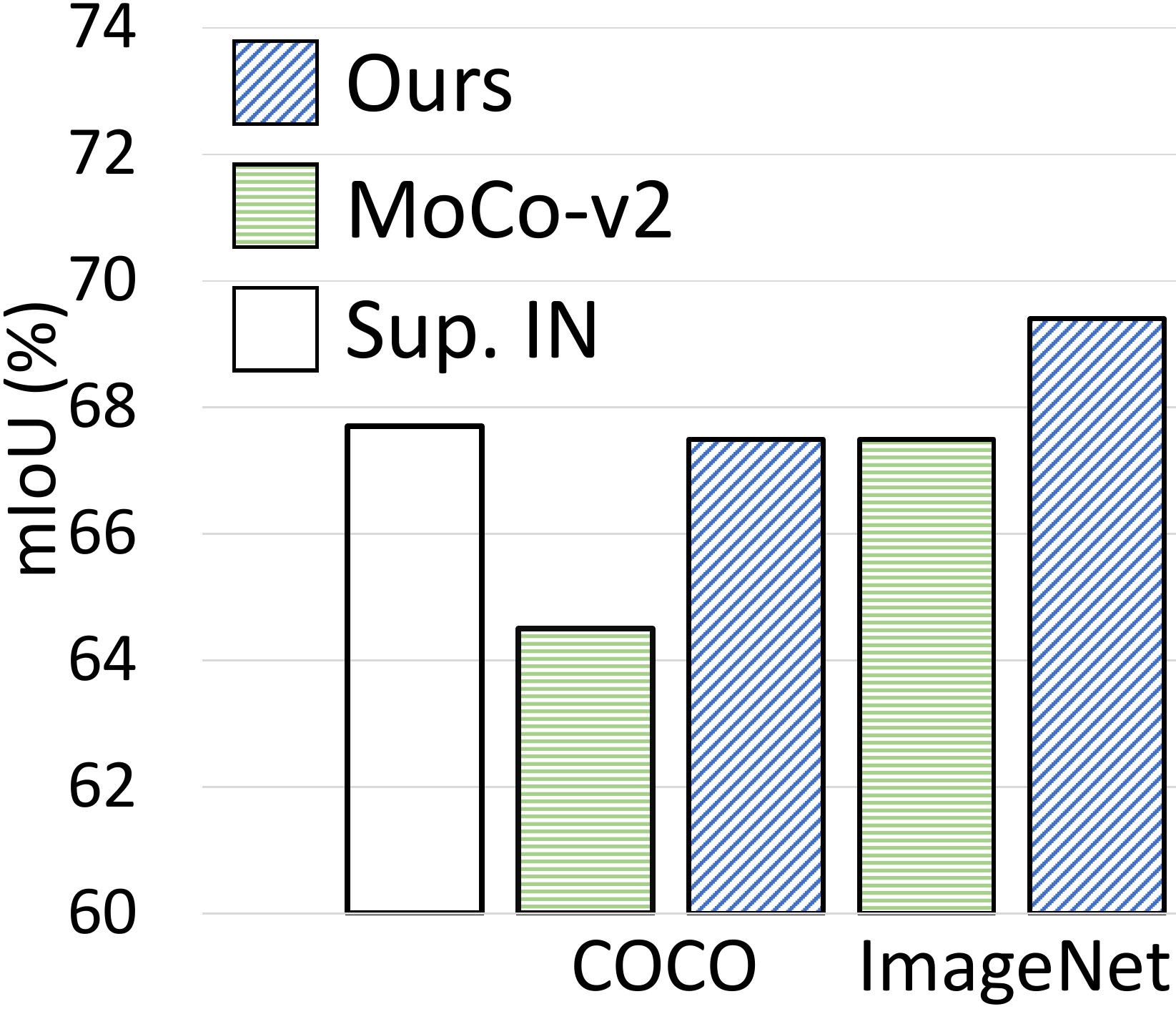}
        \label{fig:gains_sem}
    }
\caption{Comparisons of pre-trained models by fine-tuning on object detection and semantic segmentation datasets. `Sup. IN' denotes the supervised pre-training on ImageNet.
`COCO' and `ImageNet' indicate the pre-training models trained on COCO and ImageNet respectively.
(a): The object detection results of a Faster R-CNN detector fine-tuned on VOC \texttt{trainval07+12} for 24k iterations and evaluated on VOC \texttt{test2007}; (b): The semantic segmentation results of an FCN model fine-tuned on VOC \texttt{train\_aug2012} for 20k iterations and evaluated on \texttt{val2012}.
The results are averaged over 5 independent trials.
}
\label{fig:performance}
\end{figure}

\section{Introduction}

Pre-training has become a well-established paradigm in many computer vision tasks.
In a typical pre-training paradigm, models are first pre-trained on large-scale datasets and then fine-tuned on target tasks with less training data.
Specifically, the supervised ImageNet pre-training has been 
dominant for years, where the models are pre-trained to solve image classification and transferred to
downstream 
tasks.
However, there is a gap between image classification pre-training and target dense prediction tasks, such as  object detection~\cite{voc,coco} and semantic segmentation~\cite{cityscapes}.
The former focuses on assigning a category 
to 
an input image, while the latter needs to perform dense classification or regression over the whole image.
For example, semantic segmentation aims to assign a category for each pixel, and object detection aims to predict the categories and bounding boxes for all 
object instances 
of interest.
A straightforward solution 
would be 
to pre-train on dense prediction tasks directly. 
However, these tasks' annotation is notoriously time-consuming  compared to the image-level labeling, making it hard to collect data at a massive scale to pre-train a universal feature representation.

Recently, unsupervised visual pre-training has attracted much research attention, which aims to learn a proper visual representation from a large set of unlabeled images.
A few methods~\cite{moco,simclr,mocov2,byol} show the effectiveness in downstream tasks, which achieve comparable or better results compared to supervised ImageNet pre-training.
However, the gap between image classification pre-training and target dense prediction tasks still exists.
First, %
almost all recent 
self-supervised learning methods %
formulate the learning as image-level prediction using global features. 
They all can be thought of as classifying each image into its own %
version, \ie, instance discrimination~\cite{wu2018unsupervised}.
Moreover, %
existing approaches are usually evaluated and optimized on the image classification benchmark. %
Nevertheless, better image classification does not guarantee more accurate object detection, as shown in \cite{he2019rethinking}.
Thus,  self-supervised learning that is customized for dense prediction tasks is on demand. 
As for unsupervised pre-training,  dense annotation is no longer %
needed. 
A clear approach  would be pre-training as a dense prediction task \textit{directly}, thus 
removing  the gap between pre-training and target dense prediction tasks.

Inspired by the supervised dense prediction tasks, \eg, semantic segmentation, which performs dense per-pixel classification, we propose dense contrastive learning (DenseCL) for self-supervised visual pre-training.
\Ours views the self-supervised learning task as a dense %
pairwise contrastive learning 
rather than the global image classification.
First, we introduce a dense projection head that takes the features from backbone networks as input and generates dense feature vectors.
Our method naturally preserves the spatial information and constructs a dense output format, compared to the existing global projection head that applies a global pooling to the backbone features and outputs a single, global feature vector for each  image.
Second, we define the positive sample of each local feature vector by extracting the correspondence across views.
To construct an unsupervised objective function, we further design a dense contrastive loss, which extends the conventional InfoNCE loss~\cite{oord2018representation} to a dense paradigm.
With the above approaches, we perform contrastive learning densely using a fully convolutional network (FCN)~\cite{long2015fully}, 
similar 
to target dense prediction tasks.

Our main contributions are thus summarized as follows.
\begin{itemize}
\itemsep -0.051cm 
    \item
    We propose a new contrastive learning paradigm, \ie, dense contrastive learning, which performs dense pairwise contrastive learning at the level of pixels (or local features).
    \item
    With the proposed dense contrastive learning, we design a simple and effective self-supervised learning method tailored for dense prediction tasks, termed \Ours, which fills the gap between self-supervised pre-training and dense prediction tasks.
     \item
    \Ours significantly outperforms the state-of-the-art MoCo-v2~\cite{mocov2} when transferring the pre-trained model to downstream dense prediction tasks, including object detection ($+2.0\%$ AP), instance segmentation ($+0.9\%$ AP) and semantic segmentation ($+3.0\%$ mIoU), and far surpasses the supervised ImageNet pre-training.
\end{itemize}

\begin{figure*}[ht!]
    \centering
    \subfloat[Global Contrastive Learning]{
        \includegraphics[width=0.48\textwidth]{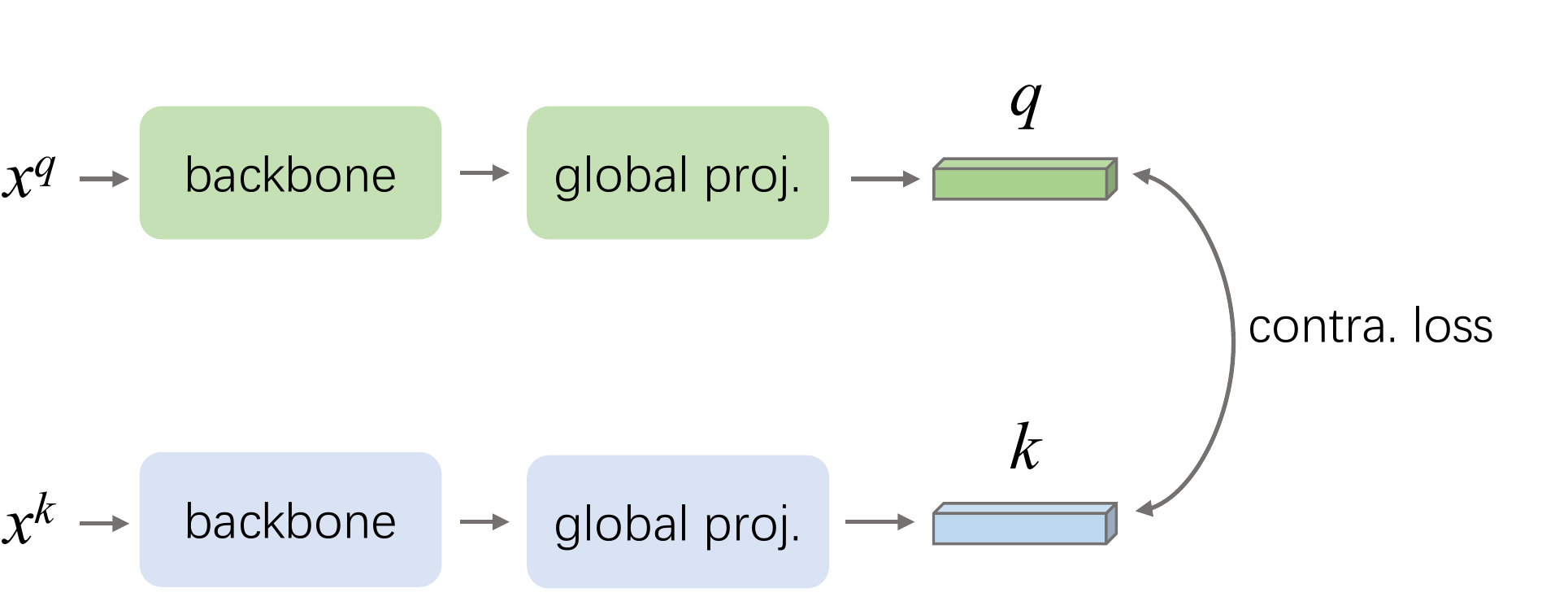}
        \label{fig:pipeline_moco}
    }
    \hspace{0.1in}
    \subfloat[Dense Contrastive Learning]{
        \includegraphics[width=0.48\textwidth]{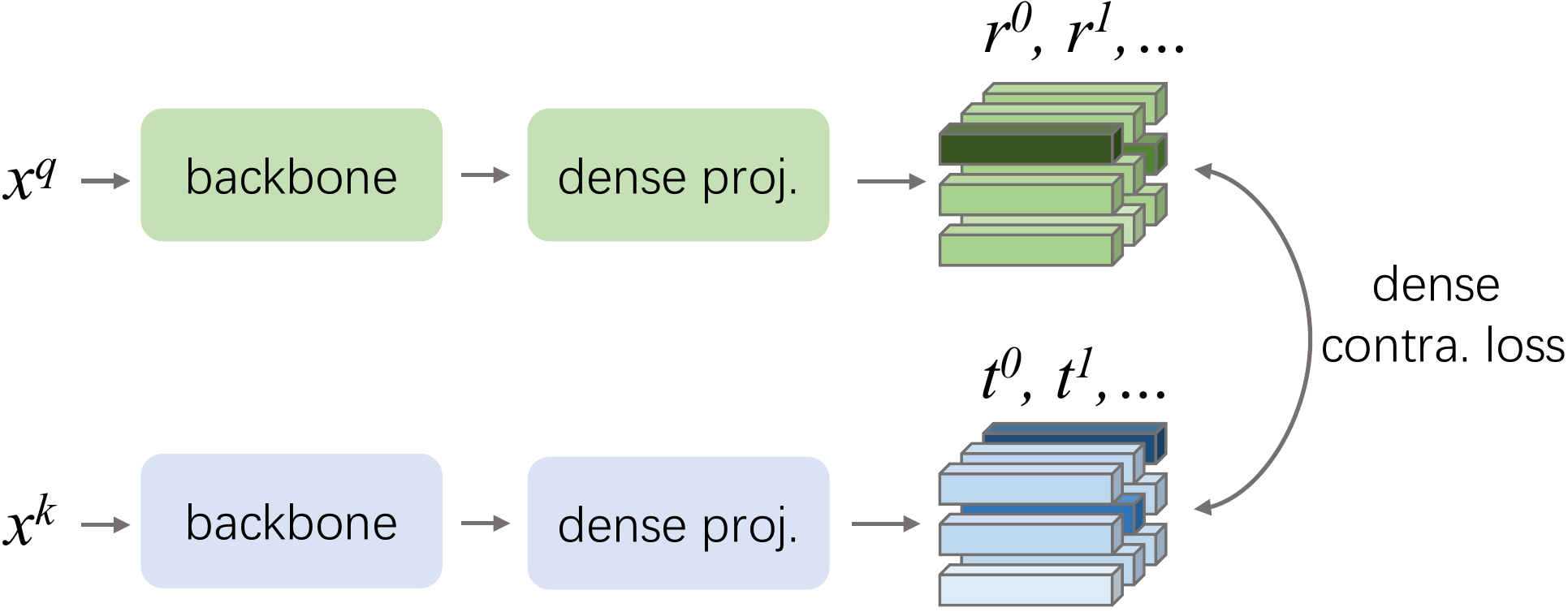}
        \label{fig:pipeline_ours}
    }
\caption{Conceptual illustration of two contrastive learning paradigms for representation learning. We use a pair of query and key for simpler illustration. 
The backbone can be any convolutional neural network. 
(a): The contrastive loss is computed between the single feature vectors outputted by the global projection head, at the level of global feature;
(b): The dense contrastive loss is computed between the dense feature vectors outputted by the dense projection head, at the level of local feature.
For both paradigms, the two branches can be the same encoder or different ones, \eg, an encoder and its momentum-updated one.
}
\vspace{-0.75em}
\label{fig:pipeline}
\end{figure*}

\subsection{Related Work}
\myparagraph{Self-supervised pre-training.} 
Generally speaking, the success of self-supervised learning~\cite{wu2018unsupervised,moco,xie2020pointcontrast,zhao2020makes,han2020self,byol} can be attributed to two important aspects namely \textit{contrastive learning}, and \textit{pretext tasks}. The objective functions used to train visual representations in many  methods are either reconstruction-based loss functions~\cite{context15,inpainting16,gan}, or contrastive loss that
measures the co-occurrence of multiple views~\cite{tian2019contrastive}.
Contrastive learning, holds the key to most state-of-the-art methods~\cite{wu2018unsupervised,moco,simclr,xie2020pointcontrast}, in which the positive pair is usually formed with two augmented views of the same image (or other visual patterns), while negative ones are formed with different images.

A wide range of pretext tasks have been %
explored 
to learn a good representation. These examples include colorization~\cite{zhang2016colorful}, context autoencoders~\cite{context15}, inpainting~\cite{inpainting16}, spatial jigsaw puzzles~\cite{jigsaw} and discriminate orientation~\cite{gidaris2018rotations}.
These methods achieved very limited success in computer vision. 
The breakthrough approach is SimCLR \cite{simclr}, which 
follows an instance discrimination pretext task, similar to~\cite{wu2018unsupervised}, where the features of each instance are pulled away from those of all other instances in the training set. 
Invariances are encoded from low-level image transformations such as cropping, scaling, and color jittering. 
Contrastive learning and pretext tasks are often combined to form a representation learning framework.
\Ours belongs to the self-supervised pre-training paradigm, 
and we naturally make the framework friendly for dense prediction tasks such as semantic segmentation and object detection.

\myparagraph{Pre-training for dense prediction tasks.}
Pre-training has enabled surprising results on many dense prediction tasks, including object detection~\cite{ren2015faster,yolo} and semantic segmentation~\cite{long2015fully}.
These models are usually fine-tuned from ImageNet pre-trained model, which is designed for image-level recognition tasks. 
Some previous studies have shown the gap between ImageNet pre-training and dense prediction tasks in the context of network architecture~\cite{li2018detnet,kong2016hypernet,efficientdet,hrnet}. 
YOLO9000~\cite{redmon2017yolo9000} proposes to joint train the object detector on both classification and detection data. 
He et al.~\cite{he2019rethinking} demonstrate that even we pre-train on extremely larger classification dataset (\eg, Instagram~\cite{mahajan2018exploring}, which is 3000$\times$ larger than ImageNet), the transfer improvements on object detection are relatively small.
Recent works~\cite{li2019analysis,zhou2020cheaperpretrain} show that
pre-trained models utilizing object detection data and annotations (\eg MS COCO~\cite{coco}) could achieve on par performance on object detection and semantic segmentation compared with ImageNet pre-trained model.
While the supervised pre-training for dense prediction tasks has been explored before \Ours, there are few works on designing an unsupervised paradigm for dense prediction tasks.
Concurrent and independent works~\cite{pinheiro2020unsupervised, chaitanya2020contrastive} also find that contrastive learning at the level of local features matters.
One of the main differences is that they construct the positive pairs according to the geometric transformation, which brings the following issues.
1) Inflexible data augmentation. Need careful design for each kind of data augmentation to maintain the dense matching.
2) Limited application scenarios. It would fail when the geometric transformation between two views are not available. For example, two images are sampled from a video clip as the positive pair, which is the case of learning representation from video stream.
By contrast, our method is totally decoupled from the data pre-processing, thus enabling fast and flexible training while being agnostic about what kind of augmentation is used and how the images are sampled.

\myparagraph{Visual correspondence.}
The visual correspondence problem is to compute the pairs of pixels from two images that result from the same scene~\cite{zabih1994correspondence}, 
and it is crucial for %
many
applications, including optical flow~\cite{dosovitskiy2015flownet}, structure-from-motion~\cite{schonberger2016structure}, visual
SLAM~\cite{kerl2013dense}, 3D reconstruction~\cite{stereoscan} \etc. 
Visual correspondence could be formulated as the problem of learning feature similarity between matched patches or points. 
Recently, a variety of convolutional neural network based approaches are proposed to measure the similarity between patches across images, including both supervised~\cite{choy2016universal,kim2017fcss} and unsupervised ones~\cite{zhang2016unsupervised,halimi2019unsupervised}.
Previous works usually leverage explicit supervision to learn the correspondence for a specific application. 
\Ours learns general representations that could be shared among multiple dense prediction tasks. 

\section{Method}

\subsection{Background}
\label{sub:background}

For self-supervised representation learning, the breakthrough approaches are 
MoCo-v1/v2~\cite{moco, mocov2} and SimCLR~\cite{simclr}, which both employ contrastive unsupervised learning to learn good representations from unlabeled data.
We briefly introduce the state-of-the-art self-supervised learning framework by abstracting a common paradigm.

\myparagraph{Pipeline.}
Given an unlabeled dataset, an instance discrimination~\cite{wu2018unsupervised} pretext task is followed where the features of each image in the training set are pulled away from those of other images.
For each image, random `views' are generated by random data augmentation.
Each view is fed into an encoder for extracting features that encode and represent the whole view.
There are two core components in an encoder, \ie, the backbone network and the projection head.
The projection head attaches to the backbone network.
The backbone is the model to be transferred after pre-training, while the projection head will be thrown away once the pre-training is completed.
For a pair of views, they can be encoded by the same encoder~\cite{simclr}, or separately by an encoder and its momentum-updated one~\cite{moco}.
The encoder is trained by optimizing a pairwise contrastive (dis)similarity loss, as revisited below.
The overall pipeline is illustrated in Figure~\ref{fig:pipeline_moco}.

\myparagraph{Loss function.}
Following the principle of MoCo~\cite{moco}, the contrastive learning can be considered as a dictionary look-up task.
For each encoded query $q$, there is a set of encoded keys $\{ k_0, k_1, ... \}$, among which a single positive key $k_{+}$ matches query $q$.
The encoded query and keys are generated from different views.
For an encoded query $q$, its positive key $k_{+}$ encode different views of the same image, while the negative keys encode the views of different images.
A contrastive loss function InfoNCE~\cite{oord2018representation} is employed to pull $q$ close to $k_{+}$ while pushing it away from other negative keys:
\begin{equation}
\small
\mathcal{L}_q = -\log \frac{\exp(q {\cdot} k_{+} / \tau)} {\exp(q {\cdot} k_{+}) + \sum_{k_-} \exp(q {\cdot} k_-  / \tau)},
\label{eq:contra_loss}
\end{equation}
where $\tau$ denotes a temperature hyper-parameter as in \cite{wu2018unsupervised}.

\subsection{\Ours Pipeline}
We propose a new self-supervised learning framework tailored for dense prediction tasks, termed \Ours.
\Ours extends and generalizes the existing framework to a dense paradigm.
Compared to the existing paradigm revisited in~\ref{sub:background}, the core differences lie in the encoder and loss function.
Given an input view, the dense feature maps are extracted by the backbone network, \eg, ResNet~\cite{he2016deep} or any other convolutional neural network, and forwarded to the following projection head.
The projection head consists of two sub-heads in parallel, which are global projection head and dense projection head respectively.
The global projection head can be instantiated as any of the existing projection heads such as the ones in~\cite{moco, simclr, mocov2}, which takes the dense feature maps as input and generates a global feature vector for each view.
For example, the projection head in~\cite{mocov2} consists of a global pooling layer and an MLP which contains two fully connected layers with a ReLU layer between them.
In contrast, the dense projection head takes the same input but outputs dense feature vectors. 

Specifically, the global pooling layer is removed and the MLP is 
replaced by the identical 1$\times$1 convolution layers~\cite{long2015fully}.
In fact, the dense projection head has the same number of parameters as the global projection head.
The backbone and two parallel projection heads are end-to-end trained by optimizing a joint pairwise contrastive (dis)similarity loss at the levels of both global features and local features.

\subsection{Dense Contrastive Learning}\label{sec:method_contrastive}
\label{sec:dense_contra_learning}

We perform dense contrastive learning by extending the original contrastive loss function to a dense paradigm.
We define a set of encoded keys $\{ t_0, t_1, ... \}$ for each encoded query $r$.
However, here each query no longer represents the whole view, but encodes a local part of a view.
Specifically, it corresponds to one of the $S_h \times S_w$ feature vectors generated by the dense projection head, where $S_h$ and $S_w$ denote the spatial size of the generated dense feature maps.
Note that $S_h$ and $S_w$ can be different, but we use $S_h = S_w = S$ for simpler illustration.
Each negative key $t_-$ is the pooled feature vector of a view from a different image.
The positive key $ t_+$ is assigned according to the extracted correspondence across views, which is one of the $S^2$ feature vectors from another view of the same image.
For now, let us assume that we can easily find the positive key $ t_+$.
A discussion is deferred to the next section.
The dense contrastive loss is defined as:
\begin{equation}
\small
\mathcal{L}_{r} = \frac{1}{S^2} 
\sum_{s}
-\log \frac{\exp(r^s {\cdot} t^s_{+} / \tau)} {\exp(r^s {\cdot} t^s_{+}) + \sum_{t^s_-} \exp(r^s {\cdot} t^s_-  / \tau)},
\label{eq:dense_contra_loss}
\end{equation}
where $r^s$ denotes the $s^{\rm th}$ out of $S^2$ encoded queries.

Overall, the total loss for our \Ours can be formulated as:
\begin{equation}
\small
\mathcal{L} = (1 - \lambda)\mathcal{L}_{q} + \lambda \mathcal{L}_{r},
\label{eq:total_loss}
\end{equation}
where $\lambda$ acts as the weight to balance the two terms. 
$\lambda$ is set to 0.5 which is validated by experiments in Section~\ref{sec:exp_ablation}.

\subsection{Dense Correspondence across Views}
\label{sec:correspondence}
\def\SM{\mathbf{\Delta}}
\def\vd{\mathbf{d}}
\def\mF{\mathbf{F}}
\def\mD{\mathbf{\Theta}}

We extract the dense correspondence between the two views of the same input image.
For each view, the backbone network extracts feature maps $\mF \in \mathbb{ R}^{H\times W\times K}$, from which the dense projection head generates dense feature vectors $ \mD  \in \mathbb{ R}^{ S_h \times  S_w \times E}$.
Note that $S_h$ and $S_w$ can be different, but we use $S_h = S_w = S$ for simpler illustration.
The correspondence is built between the dense feature vectors from the two views, \ie, $\mD_1$ and $ \mD_2$. 
We match $\mD_1$ and $ \mD_2$ using the backbone feature maps $\mF_1$ and $\mF_2$.
The $\mF_1$ and $\mF_2$ are first downsampled to have the spatial shape of $S \times S$ by an adaptive average pooling, and then used to calculate the cosine similarity matrix $\SM \in \mathbb{ R}^{S^2 \times S^2}$.
The matching rule is that each feature vector in a view is matched to the most similar feature vector in another view.
Specifically, for all the $S^2$ feature vectors of $\mD_1$, the correspondence with $\mD_2$ is obtained by applying an \texttt{argmax} operation to the similarity matrix $\SM$ along the last dimension.
The matching process can be formulated as:
\begin{equation}
c_i = \mathop{\arg\max}_{j} sim(\bm f_i,\bm f^\prime_j),
\label{eq:match}
\end{equation}
where  
$\bm f_i$ is the $i^{\rm th}$ feature vector of backbone feature maps $\mF_1$, and $\bm f^\prime_j$ is the $j^{\rm th}$  of $\mF_2$.
${sim}(\bm u,\bm v)$ denotes the cosine similarity, calculated by the dot product between $\ell_2$ normalized $\bm u$ and $\bm v$, \ie, ${sim}(\bm u,\bm v) = \bm u^\top \bm v / \lVert\bm u\rVert \lVert\bm v\rVert$.
The obtained $c_i$ denotes the $i^{\rm th}$ out of $S^2$ matching from $\mD_1$ to $\mD_2$, which means that $i^{\rm th}$ feature vector of $\mD_1$ matches ${c_i}^{\rm th}$ of $\mD_2$.
The whole matching process could be efficiently implemented by matrix
operations, thus introducing 
negligible
latency overhead.

For the simplest case where $S = 1$, the matching degenerates into the one in global contrastive learning as the  single correspondence naturally exists between two global feature vectors, which is the case introduced in Section~\ref{sub:background}.

According to the extracted dense correspondence, one can easily find the positive key $t_+$ for each query $r$ during the dense contrastive learning introduced in Section~\ref{sec:dense_contra_learning}. 

Note that without the global contrastive learning term (\ie, $\lambda=1$), there is a chicken-and-egg 
issue that good features will not be learned if incorrect 
correspondence is extracted, and the correct correspondence will not be
available  if the features are not sufficiently good.
In our default setting where $\lambda=0.5$, no unstable
training is observed.
Besides setting $\lambda \in (0, 1)$ during the whole training, we introduce two more solutions which can also tackle this problem, detailed in Section~\ref{sec:behavior}.

\section{Experiments}

We adopt MoCo-v2~\cite{mocov2} as our baseline method, as which shows the state-of-the-art results and outperforms other methods by a large margin on downstream object detection task, as shown in Table~\ref{tab:voc}.
It indicates that it should serve as a very strong baseline on which we can demonstrate the effectiveness of our approach.

\myparagraph{Technical details.}
We adapt most of the settings from~\cite{mocov2}.
A ResNet~\cite{he2016deep} is adopted as the backbone. 
The following global projection head and dense projection head both have a fixed-dimensional output. The former outputs a single 128-D feature vector for each input and the latter outputs dense 128-D feature vectors.
Specifically, the dense projection head consists of adaptive average pooling (optional), $1 \times 1$ convolution, ReLU, and $1 \times 1$ convolution.
Following~\cite{simclr, mocov2}, the hidden layer's dimension is 2048, and the final output dimension is 128.
Each $\ell_2$ normalized feature vector represents a query or key.
For both the global and dense contrastive learning, the dictionary size is set to 65536.
The momentum is set to 0.999.
Shuffling BN~\cite{moco} is used during the training.
The temperature $\tau$ in Equation~(\ref{eq:contra_loss}) and Equation~(\ref{eq:dense_contra_loss}) is set to 0.2.
The data augmentation pipeline consists of $224\times224$-pixel ramdom resized cropping, random color jittering, random gray-scale conversion, gaussian blurring and random horizontal flip.

\subsection{Experimental Settings}
\label{sec:exp_settings}

\myparagraph{Datasets.}
The pre-training experiments are conducted on two large-scale datasets:  MS COCO~\cite{coco} and ImageNet~\cite{imagenet}.
Only the training sets are used during the pre-training, which are $\app$118k and $\app$1.28 million images respectively.
COCO and ImageNet represent two kinds of image data.
The former is more natural and real-world, containing diverse scenes in the wild.
It is a widely used and challenging dataset for object-level and pixel-level recognition tasks, 
such as object detection and instance segmentation.
While the latter is heavily curated, carefully constructed for image-level recognition.
A clear and quantitative comparison is the number of objects of interest.
For example, COCO has a total of 123k images and 896k labeled objects, an average of 7.3 objects per image, which is far more than the ImageNet DET dataset's 1.1 objects per image.

\myparagraph{Pre-training setup.}
For ImageNet pre-training, we closely follow MoCo-v2~\cite{mocov2} and use the same training hyper-parameters.
For COCO pre-training including both baseline and ours, we use an initial learning rate of 0.3 instead of the original 0.03, 
as the former shows better performance in MoCo-v2 baseline when pre-training on COCO.
We adopt SGD as the optimizer and we set its weight decay and momentum to 0.0001 and 0.9.
Each pre-training model is optimized on 8 GPUs with a cosine learning rate decay schedule and a mini-batch size of 256.
We train for 800 epochs for COCO, which is a total $\app$368k iterations.
For ImageNet, we train for 200 epochs, a total of 1 million  iterations.

\myparagraph{Evaluation protocol.}
We evaluate the pre-trained models by fine-tuning on the target dense prediction tasks end-to-end.
Challenging and popular datasets are adopted to fine-tune mainstream algorithms for different target tasks, \ie VOC object detection, COCO object detection, COCO instance segmentation, VOC semantic segmentation, and Cityscapes semantic segmentation.
When evaluating on object detection, we follow the common protocol that fine-tuning a Faster R-CNN detector (C4-backbone) on the VOC  \texttt{trainval07+12} set with standard 2x schedule in~\cite{wu2019detectron2} and testing on the VOC  \texttt{test2007} set.

In addition, we evaluate object detection and instance segmentation by fine-tuning a Mask R-CNN detector (FPN-backbone) with on COCO \texttt{train2017} split ($\app$118k images) with the standard 
1$\times$ 
schedule and evaluating on COCO 5k \texttt{val2017} split.
we follow the settings in~\cite{tian2020makes}.
Synchronized batch normalization is used in backbone, FPN~\cite{fpn} and prediction heads during the training.

For semantic segmentation, an FCN model~\cite{long2015fully} is fine-tuned on VOC \texttt{train\_aug2012} set (10582 images) for 20k iterations and evaluated on \texttt{val2012} set.
We also evaluate semantic segmentation on Cityscapes dataset by training an FCN model on \texttt{train\_fine} set (2975 images) for 40k iterations and test on \texttt{val} set.
We follow the settings in \texttt{mmsegmentation}~\cite{mmseg}, except that the first $7 \times 7$ convolution is kept to be consistent with the pre-trained models. 
Batch size is set to 16. Synchronized batch normalization is used. Crop size is 512 for VOC~\cite{voc} and 769 for Cityscapes~\cite{cityscapes}.

\subsection{Main Results}

\myparagraph{PASCAL VOC object detection.}
In Table~\ref{tab:voc}, we report the object detection result on PASCAL VOC and compare it with other state-of-the-art methods.
When pre-trained on COCO, our \Ours outperforms the MoCo-v2 baseline by 2\% AP.
When pre-trained on ImageNet, the MoCo-v2 baseline has already surpassed other state-of-the-art self-supervised learning methods.
And \Ours still yields  1.7\% AP improvements, strongly demonstrating the effectiveness of our method.
The gains are consistent over all three metrics.
It should be noted that we achieve much larger improvements on more stringent AP$_{75}$ compared to those on AP$_{50}$, which indicates \Ours largely helps improve the localization accuracy.
Compared to the supervised ImageNet pre-training, we achieve the significant 4.5\% AP gains.

\begin{table}[t]
\small
\centering
\begin{tabular}{l|c|ccc}
pre-train & AP & AP$_\text{50}$ & AP$_\text{75}$ \\ 
\hline
random init. & 32.8 & 59.0 & 31.6 \\
super. IN &  54.2 & 81.6 & 59.8 \\
\hline
MoCo-v2 CC & 54.7 & 81.0 & 60.6 \\
\textbf{\Ours} CC &  56.7 &  81.7 & 63.0 \\
\hline
SimCLR IN~\cite{simclr} & 51.5 & 79.4 & 55.6 \\
BYOL IN~\cite{byol} & 51.9 & 81.0 & 56.5 \\
MoCo IN~\cite{moco} & 55.9 & 81.5 & 62.6 \\
MoCo-v2 IN~\cite{mocov2} & 57.0 & 82.4 & 63.6 \\
MoCo-v2 IN* &  57.0 & 82.2 & 63.4 \\
\textbf{\Ours} IN &  58.7 & 82.8 & 65.2\\

\end{tabular}
\caption{\textbf{Object detection fine-tuned on PASCAL VOC.} 
`CC' and `IN' indicate the pre-training models trained on COCO and ImageNet respectively.
The models pre-trained on the same dataset are with the same training epochs, \ie,
800 epochs for COCO and 200 epochs for ImageNet.
`*' means re-implementation.
The results of other methods are either from their papers or third-party implementation.
All the detectors are trained on \texttt{trainval07+12} for 24k iterations and evaluated on \texttt{test2007}. 
The metrics include the VOC metric AP$_{50}$ (\ie, IoU threshold is 50\%) and COCO-style AP and AP$_{75}$.
The results are averaged over 5 independent trials.
}
\label{tab:voc}
\end{table}

\myparagraph{COCO object detection and segmentation.}
The object detection and instance segmentation results on COCO are reported in Table~\ref{tab:coco}.
For object detection, \Ours outperforms MoCo-v2 by 1.1\% AP and 0.5\% AP when pre-trained on COCO and ImageNet respectively.
The gains are 0.9\% AP and 0.3\% AP for instance segmentation.
Note that fine-tuning on COCO with a COCO pre-trained model is not a typical scenario.
But the clear improvements still show the effectiveness.

In Table~\ref{tab:semi_sup_det}, we further evaluate the pre-trained models on semi-supervised object detection.
In this semi-supervised setting, only 10\% training data is used during the fine-tuning.
\Ours outperforms MoCo-v2 by 1.3\% AP$^{\rm b}$ and 1.0\% AP$^{\rm b}$ when pre-training on COCO and ImageNet respectively.
It should be noted that the gains are more significant than that of the fully-supervised setting which uses all of $\app$118k images during the fine-tuning.
For example, when pre-training on ImageNet, \Ours surpasses MoCo-v2 by 1.0\% AP$^{\rm b}$ and 0.5\% AP$^{\rm b}$ for semi-supervised setting and fully-supervised setting respectively.

\begin{table}[t]
\small
\centering
\small
\tablestyle{5pt}{1.1}
\begin{tabular}{ l |c|cc|c|cc}
pre-train & AP$^{\rm b}$ & AP$^{\rm b}_{50}$ & AP$^{\rm b}_{75}$ & AP$^{\rm m}$ & AP$^{\rm m}_{50}$ & AP$^{\rm m}_{75}$ \\
\hline
random init. & 32.8 & 50.9 & 35.3 & 29.9 & 47.9 & 32.0\\
super. IN & 39.7 & 59.5 & 43.3 & 35.9 & 56.6 & 38.6 \\
\hline
MoCo-v2 CC & 38.5 & 58.1 & 42.1 & 34.8 & 55.3 & 37.3 \\
\textbf{\Ours} CC & 39.6 & 59.3 & 43.3 & 35.7 & 56.5 & 38.4 \\
\hline
SimCLR IN & 38.5 & 58.0 & 42.0 & 34.8 & 55.2 & 37.2 \\
BYOL IN & 38.4& 57.9& 41.9 & 34.9 & 55.3 & 37.5\\
MoCo-v2 IN & 39.8 & 59.8 & 43.6 & 36.1 & 56.9 & 38.7\\
\textbf{\Ours} IN & 40.3 & 59.9 & 44.3 & 36.4 & 57.0 & 39.2 \\
\end{tabular}
\caption{\textbf{Object detection and instance segmentation fine-tuned on COCO.} 
`CC' and `IN' indicate the pre-training models trained on COCO and ImageNet respectively.
All the detectors are trained on \texttt{train2017} with default $1\times$ schedule and evaluated on \texttt{val2017}. 
The metrics include bounding box AP (AP$^{\rm b}$) and mask AP (AP$^{\rm m}$).
}
\label{tab:coco}
\end{table}

\begin{table}[t!]
\small
\centering
\tablestyle{5pt}{1.1}
\begin{tabular}{ l |c|cc|c|cc}
pre-train & AP$^{\rm b}$ & AP$^{\rm b}_{50}$ & AP$^{\rm b}_{75}$ & AP$^{\rm m}$ & AP$^{\rm m}_{50}$ & AP$^{\rm m}_{75}$ \\
\hline
random init. & 20.6 & 34.0 & 21.5 & 18.9 & 31.7 & 19.8 \\
super. IN & 23.6 & 37.7 & 25.4 & 21.8 & 35.4 & 23.2\\
\hline
MoCo-v2 CC & 22.8 & 36.4 & 24.2 & 20.9 & 34.6 & 21.9 \\
\textbf{\Ours} CC & 24.1 & 38.1 & 25.6 & 21.9 & 36.0 & 23.0 \\
\hline
MoCo-v2 IN & 23.8 & 37.5 & 25.6 & 21.8 & 35.4 & 23.2\\
\textbf{\Ours} IN & 24.8 & 38.8 & 26.8 & 22.6 & 36.8 & 23.9 \\
\end{tabular}
\caption{\textbf{Semi-supervised object detection and instance segmentation fine-tuned on COCO.}
During the fine-tuning, only 10\% training data is used.
`CC' and `IN' indicate the pre-training models trained on COCO and ImageNet respectively.
All the detectors are trained on \texttt{train2017} for 90k iterations and evaluated on \texttt{val2017}. 
The metrics include bounding box AP (AP$^{\rm b}$) and mask AP (AP$^{\rm m}$).
}
\label{tab:semi_sup_det}
\end{table}

\myparagraph{PASCAL VOC semantic segmentation.}
We show the largest improvements on semantic segmentation.
As shown in Table~\ref{tab:seg}, \Ours yields 3\% mIoU gains when pre-training on COCO and fine-tuning an FCN on VOC semantic segmentation.
The COCO pre-trained \Ours achieves the same 67.5\% mIoU as ImageNet pre-trained MoCo-v2.
Note that compared to 200-epoch ImageNet pre-training, 800-epoch COCO pre-training only uses $\app1/10$ images and $\app1/3$ iterations.
When pre-trained on ImageNet, \Ours consistently brings 1.9\% mIoU gains.
It should be noted that the ImageNet pre-trained MoCo-v2 
shows no transfer superiority compared with
the supervised counterpart (67.5\% vs.\  67.7\% mIoU).
But \Ours outperforms the supervised pre-training by a large margin, \ie, 1.7\% mIoU.

\myparagraph{Cityscapes semantic segmentation.}
Cityscapes is a benchmark largely different from the above VOC and COCO.
It focuses on urban street scenes.
Nevertheless, in Table~\ref{tab:seg}, we observe the same performance boost with \Ours.
Even the COCO pre-trained \Ours can surpass the supervised ImageNet pre-trained model by 1.9\% mIoU.

\begin{table}
\setlength{\tabcolsep}{5pt}
\centering 
\small
\subfloat[PASCAL VOC]{
\begin{tabular}{l|c}
pre-train & mIoU \\ 
\hline
random init. & 40.7 \\
super. IN &  67.7 \\
\hline
MoCo-v2 CC & 64.5 \\
\textbf{\Ours} CC & 67.5\\
\hline
SimCLR IN &  64.3 \\
BYOL IN & 63.3 \\
MoCo-v2 IN &  67.5\\
\textbf{\Ours} IN &  69.4   \\

\end{tabular}
}
\hspace{0.5em}
\subfloat[Cityscapes]{
\begin{tabular}{l|c}
pre-train & mIoU \\ 
\hline
random init. & 63.5 \\
super. IN &  73.7 \\
\hline
MoCo-v2 CC & 73.8 \\
\textbf{\Ours} CC & 75.6  \\
\hline
SimCLR IN &  73.1 \\
BYOL IN & 71.6 \\
MoCo-v2 IN & 74.5 \\
\textbf{\Ours} IN &  75.7  \\

\end{tabular}
}
\caption{\textbf{Semantic segmentation on PASCAL VOC and Cityscapes.} 
`CC' and `IN' indicate the pre-training models trained on COCO and ImageNet respectively.
The metric is the commonly used mean IoU (mIoU).
Results are averaged over 5 independent trials.
}
\label{tab:seg}
\vspace{-1em}
\end{table}

\subsection{Ablation Study}
\label{sec:exp_ablation}

We conduct extensive ablation experiments to show how each component contributes to \Ours.
We report ablation studies by pre-training on COCO and fine-tuning on VOC0712 object detection, as introduced in  Section~\ref{sec:exp_settings}.
All the detection results are averaged over 5 independent trials.
We also provide results of VOC2007 SVM Classification, following~\cite{GoyalM0M19, ZhanX0OL20} which train linear SVMs on the VOC \texttt{train2007} split using the features extracted from the frozen backbone  and evaluate on the \texttt{test2007} split.

\myparagraph{Loss weight $\lambda$.}
The hyper-parameter $\lambda$ in Equation~(\ref{eq:total_loss}) serves as the weight to balance the two contrastive loss terms, \ie, the global term and the dense term.
We report the results of different $\lambda$ in Table~\ref{tab:ablation_weight}.
It shows a trend that the detection performance improves when we increase the $\lambda$.
For the baseline method, \ie, $\lambda=0$, the result is 54.7\% AP.
The AP is 56.2\% when $\lambda=0.3$, which improves the baseline by 1.5\% AP.
Increasing $\lambda$ from 0.3 to 0.5 brings another 0.5\% AP gains.
Although further increasing it to 0.7 still gives minor improvements (0.1\% AP) on detection performance, the classification result drops from 82.9\% to 81.0\%. 
Considering the trade-off, we use $\lambda=0.5$ as our default setting in other experiments.
It should be noted that when $\lambda = 0.9$, compared to the MoCo-v2 baseline, the classification performance rapidly drops (-4.8\% mAP) while the detection performance improves for 0.8\% AP.
It is in accordance with our intention that \Ours is specifically designed for dense prediction tasks.

\begin{table}[th]
\small
\centering
\begin{tabular}{c|ccc|c}
& \multicolumn{3}{c}{Detection} & 
\multicolumn{1}{|c}{Classification} \\
$\lambda$ &
  AP & AP$_\text{50}$ & AP$_\text{75}$ & mAP \\ 
\hline
0.0 & 54.7 & 81.0 & 60.6 & 82.6 \\
0.1 & 55.2 & 81.4 & 61.4 & 82.9 \\
0.3 & 56.2 & 81.5 & 62.6 & 83.3 \\
0.5 &  56.7 & 81.7 & 63.0 & 82.9 \\
0.7 & 56.8 & 81.9 & 63.1 & 81.0 \\
0.9 & 55.5 & 80.9 & 61.3 & 77.8 \\
1.0* & 53.5 & 79.5 & 58.8 & 68.9 \\
\end{tabular}
\caption{\textbf{Ablation study of weight $\lambda$.} 
$\lambda = 0$ is the MoCo-v2 baseline.
$\lambda = 0.5$ shows the best trade-off between detection and classification.
`*' indicates training with warm-up, as discussed in Section~\ref{sec:behavior}.
}
\label{tab:ablation_weight}
\end{table}

\begin{table}[th]
\small
\centering
\begin{tabular}{ l |ccc|c}
& \multicolumn{3}{c}{Detection} & 
\multicolumn{1}{|c}{Classification} \\
strategy &
  AP & AP$_\text{50}$ & AP$_\text{75}$ & mAP \\ 
\hline
random &  56.0 & 81.3 & 62.0  & 81.7 \\
max-sim $\mD$ & 56.0 & 81.5 & 62.1 & 81.8  \\
max-sim $\mF$ &  56.7 & 81.7 & 63.0 & 82.9 \\
\end{tabular}
\caption{\textbf{Ablation study of matching strategy.}
To extract the dense correspondence according to the backbone features $\mF_1$ and $\mF_2$ shows the best results.
}
\label{tab:ablation_matching}
\end{table}

\myparagraph{Matching strategy.}
In Table~\ref{tab:ablation_matching}, we compare three different 
matching strategies used to extract correspondence across views.
1) `random': the dense feature vectors from two views are randomly matched;
2) `max-sim $\mD$': the dense correspondence is extracted using the dense feature vectors $\mD_1$ and $\mD_2$ generated by the dense projection head;
(3) `max-sim $\mF$': the dense correspondence is extracted according to the backbone features $\mF_1$ and $\mF_2$, as in Equation~\ref{eq:match}.
The random matching strategy can also achieve 1.3\% AP improvements compared to MoCo-v2, meanwhile the classification performance drops by 0.9\% mAP. 
It may be because 1) the dense output format itself helps, and 2) part of the random matches are somewhat correct.
Matching by the outputs of dense projection head, \ie, $\mD_1$ and $\mD_2$, brings no clear improvement.
The best results are obtained by extracting the dense correspondence according to the backbone features $\mF_1$ and $\mF_2$.

\myparagraph{Grid size.}
In the default setting, the adopted ResNet backbone outputs features with stride 32.
For a $224\times224$-pixel crop, the backbone features $\mF$ has the spatial size of $7\times7$.
We set the spatial size of the dense feature vectors $\mD$ to $7\times7$ by default, \ie, $S = 7$.
However, $S$ can be flexibly adjusted and $\mF$ will be pooled to the designated spatial size by an adaptive average pooling, as introduced in Section~\ref{sec:correspondence}.
We report the results of using different numbers of grid in Table~\ref{tab:ablation_grid}.
For $S=1$, it is the same as the MoCo-v2 baseline except for two differences.
1) The parameters of dense projection head are independent with those of global projection head.
2) The dense contrastive learning maintains an independent dictionary.
The results are similar to those of MoCo-v2 baseline.
It indicates that the extra parameters and dictionary do not bring improvements.
The performance improves as the grid size increases.
We use grid size being 7 as the default setting, as the performance becomes stable when the $S$ grows beyond 7.

\begin{table}[th]
\small
\centering
\begin{tabular}{c|ccc|c}
& \multicolumn{3}{c}{Detection} & 
\multicolumn{1}{|c}{Classification} \\
grid size &
  AP & AP$_\text{50}$ & AP$_\text{75}$ & mAP \\ 
\hline
1 & 54.6 & 80.8 & 60.5 & 82.2 \\
3 &  55.6 & 81.3 & 61.5 & 81.6\\
5 &  56.1 & 81.4 & 62.2 & 82.6\\
7 & 56.7 & 81.7 & 63.0 & 82.9 \\
9 & 56.7 & 82.1 & 63.2 & 82.9 \\
\end{tabular}
\caption{\textbf{Ablation study of grid size $S$.} 
The results increase as the $S$ gets larger.
We use grid size being 7 in other experiments, 
as the performance becomes stable when the $S$ grows beyond 7.
}
\label{tab:ablation_grid}
\end{table}

\myparagraph{Negative samples.}
We use the global average
pooled features as negatives because it’s conceptually simpler.
Besides pooling, sampling is an alternative strategy.
For keeping the same number of negatives, one can randomly
sample a local feature from a different image. The
COCO pre-trained model with sampling strategy achieves
56.7\% AP on VOC detection, which is the same as the
adopted pooling strategy.

\begin{figure}[t!]
\centering
\includegraphics[width=0.795\linewidth]{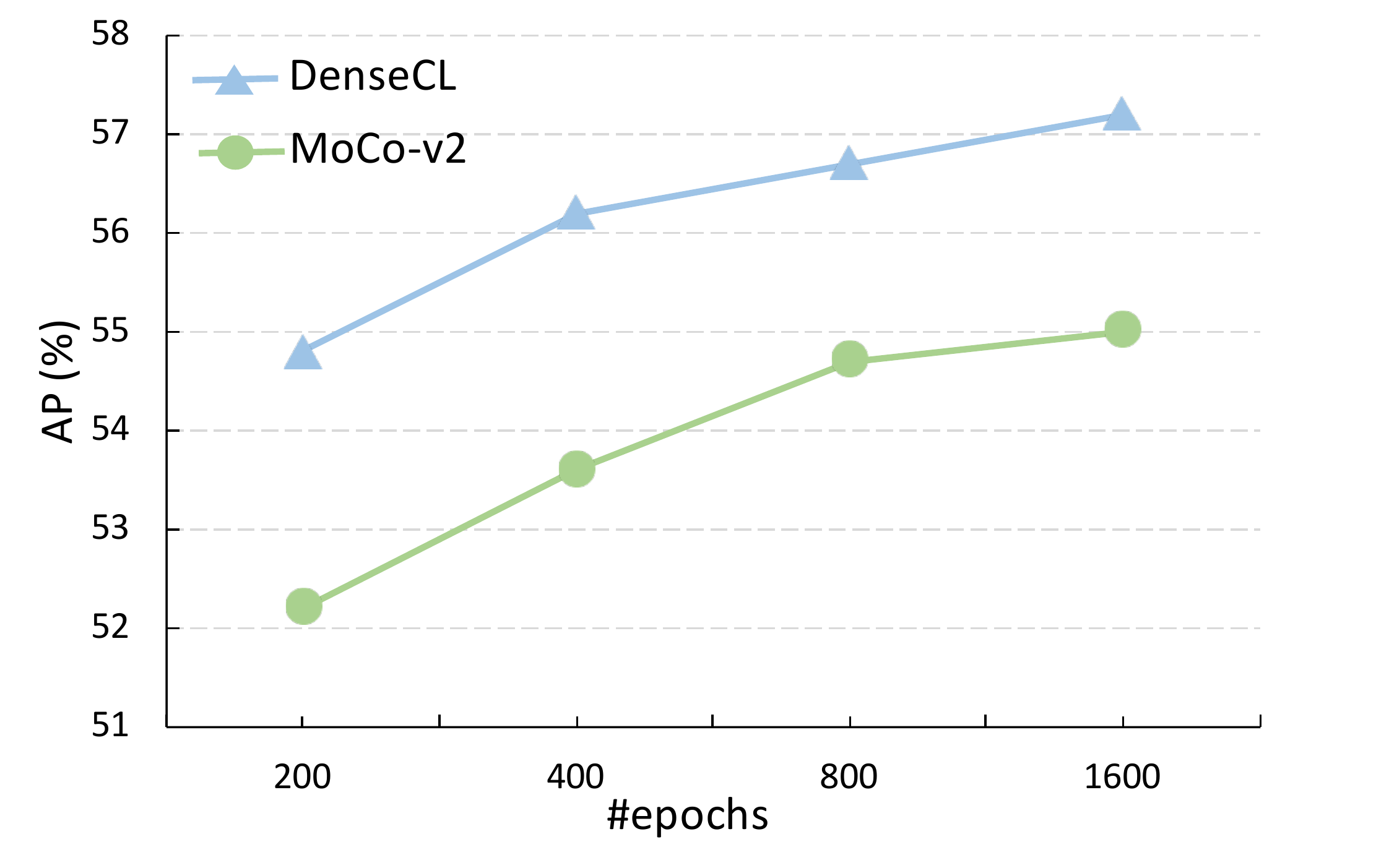}
\vspace{-0.64em}
\caption{Different pre-training schedules on COCO.
For each pre-trained model, a Faster R-CNN detector is fine-tuned on VOC \texttt{trainval07+12} for 24k iterations and evaluated on \texttt{test2007}. 
The metric is the COCO-style AP.
Results are averaged over 5 independent trials.
}
\label{fig:different_epochs}
 \vspace{-1.5em}
\end{figure}

\iftrue
\begin{figure*}[t!]
\centering
\includegraphics[width=0.95\linewidth]{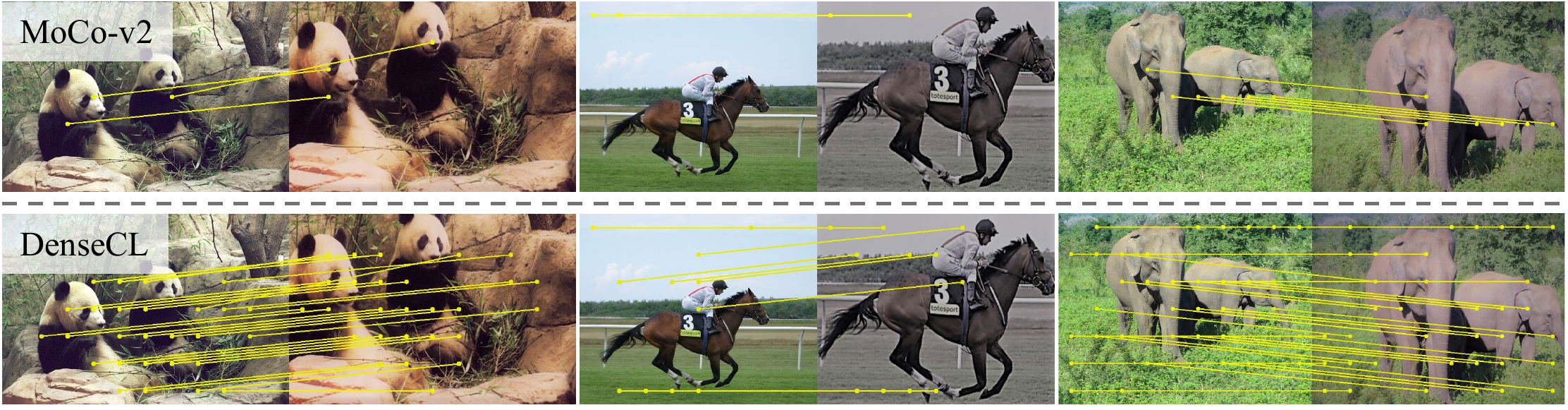}
   \caption{Visualization of dense correspondence. The correspondence is extracted between two views of the same image, using the 200-epoch ImageNet pre-trained model. \Ours extracts more high-similarity matches compared with MoCo-v2.
   Best viewed on screen.}
\label{fig:vis_correspondence_comparison}
\vspace{-0.2em}
\end{figure*}
\fi

\begin{figure*}[h!]
\centering
\includegraphics[width=0.95\linewidth]{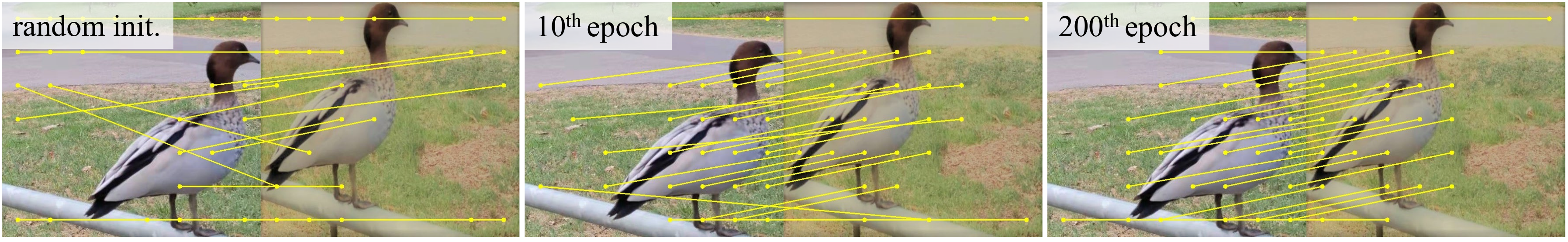}
\vspace{-0.5em}
   \caption{Comparison of dense correspondence extracted from random initialization to well trained \Ours. The correspondence is extracted between two views of the same image, using the ImageNet pre-trained model. All the matches are visualized without thresholding.}
\label{fig:correspondence}
\vspace{-0.5em}
\end{figure*}

\myparagraph{Training schedule.}
We show the results of using different training schedules in Table~\ref{tab:ablation_epoch}.
The performance consistently improves as the training schedule gets longer, from 200 epochs to 1600 epochs.
Note that the 1600-epoch COCO pre-trained \Ours even surpasses the 200-epoch ImageNet pre-trained MoCO-v2 (57.2\% AP vs. 57.0\% AP).
Compared to 200-epoch ImageNet pre-training, 1600-epoch COCO pre-training only uses $\app1/10$ images and $\app7/10$ iterations.
In Figure~\ref{fig:different_epochs}, we further provide an intuitive comparison with the baseline as the training schedule gets longer.
It shows that \Ours consistently outperforms the MoCo-v2 by at least 2\% AP.

\begin{table}[th]
\small
\centering
\begin{tabular}{c|ccc|c}
& \multicolumn{3}{c}{Detection} & 
\multicolumn{1}{|c}{Classification} \\
\#epochs &
  AP & AP$_\text{50}$ & AP$_\text{75}$ & mAP \\ 
\hline
200 & 54.8 & 80.5 & 60.7 & 77.6 \\
400 &  56.2 & 81.5 & 62.3 & 81.3 \\
800 & 56.7 & 81.7 & 63.0 & 82.9 \\
1600 & 57.2 & 82.2 & 63.6 & 83.0 \\
\end{tabular}
\caption{\textbf{Ablation study of training schedule.} 
The results consistently improve as the training schedule gets longer.
Although 1600-epoch training schedule is 0.5\% AP better, we use 800-epoch schedule in other experiments for faster training.
}
\label{tab:ablation_epoch}
\end{table}

\myparagraph{Pre-training time.}
In Table~\ref{tab:time}, we compare \Ours with MoCo-v2 in terms of training time.
\Ours is only 1s and 6s slower per epoch when pre-trained on COCO and ImageNet respectively.
The overhead is less than 1\%.
It strongly demonstrates the efficiency of our method.

\begin{table}[t]
\small
\centering
\begin{tabular}{c|cc}
time/epoch & COCO & ImageNet \\
 \hline
MoCo-v2 & $1^\prime45^{\prime\prime}$ & $16^\prime48^{\prime\prime}$ \\
\Ours & $1^\prime46^{\prime\prime}$ & $16^\prime54^{\prime\prime}$ \\
\end{tabular}
\caption{\textbf{Pre-training time comparison.} The training time per epoch is reported.
We measure the results on the same 8-GPU machine.
The training time overhead introduced by \Ours is less than 1\%.
}
\label{tab:time}
\end{table}

\subsection{
Discussions on \Ours
}
\label{sec:behavior}

To further study how \Ours works, 
in this section, 
we visualize the learned dense correspondence in \Ours. 
The issue of chicken-and-egg during the training is also discussed.

\myparagraph{Dense correspondence visualization.}
We visualize the dense correspondence from two aspects: comparison of the final correspondence extracted from different pre-training methods, \ie, MoCo-v2 vs. \Ours, and the comparison of different training status, \ie, from the random initialization to well trained \Ours.
Given two views of the same image, we use the pre-trained backbone to extract the features $\mF_1$ and $\mF_2$.
For each feature vector in $\mF_1$, we find the corresponding feature vector in $\mF_2$ which has the highest cosine similarity.
The match is kept if the same match holds from $\mF_2$ to $\mF_1$.
Each match is assigned an averaged similarity.
In Figure~\ref{fig:vis_correspondence_comparison}, we visualize the high-similarity matches (\ie, similarity $\geq 0.9$).
\Ours extracts many more high-similarity matches than its baseline.
It is in accordance with our intention that the local features extracted from the two views of the same image should be similar.

Figure~\ref{fig:correspondence} shows how the correspondence changes over training time.
The randomly initialized model extracts some random noisy matches.
The matches get more accurate as the training time increases.

\myparagraph{Chicken-and-egg issue.}
In
our pilot 
experiments, we observe that the training loss does not converge if we set $\lambda$ to 1.0, \ie, removing the global contrastive learning, and only applying
the dense contrastive learning. 
It may be because 
at the beginning of the training, 
the randomly initialized model is not able to generate correct correspondence across views.
It is thus a chicken-and-egg %
issue 
that good features will not be learned if %
incorrect 
correspondence is extracted, and the correct correspondence will not be
available 
if the features are not
sufficiently 
good.
As shown in Figure~\ref{fig:correspondence}, most of the matches are 
incorrect 
with the random initialization.
The core solution 
is
to provide a guide when training starts, to break the deadlock.
We introduce three different solutions to tackle this problem.
1) To initialize the model with the weights of a %
pre-trained model;
2) To set a warm-up period at the beginning during which the $\lambda$ is set to 0;
3) To set $\lambda \in (0, 1)$ during the whole training.
They all solve this issue well.
The second one is reported in Table~\ref{tab:ablation_weight}, with $\lambda$ being changed from 0 to 1.0 after the first 10k iterations.
We adopt the last one as the default setting 
for its simplicity.

\section{Conclusion}
In this work we have developed a simple and effective self-supervised learning framework \Ours, which is designed and optimized for dense prediction tasks.
A new contrastive learning paradigm is proposed to perform dense pairwise contrastive learning at the level of pixels (or local features).
Our method largely closes the gap between self-supervised pre-training and dense prediction tasks, and shows significant improvements in a variety of tasks and datasets, including PASCAL VOC object detection, COCO object detection, COCO instance segmentation, PASCAL VOC semantic segmentation and Cityscapes semantic segmentation.
We expect the proposed effective and efficient self-supervised pre-training techniques could be applied to larger-scale data to fully realize its potential, as well as hoping that \Ours pre-trained models would completely replace the supervised pre-trained models in many of those dense prediction tasks in computer vision.

\textbf{Acknowledgement and Declaration of Conflicting Interests}
The authors would like to thank Wanxuan Lu for her helpful discussion.
CS and his employer received no financial
support for the research, authorship, and/or publication of this article.

{\small
\bibliographystyle{ieee_fullname}
\bibliography{draft}

\begin{thebibliography}{10}\itemsep=-1pt

\bibitem{chaitanya2020contrastive}
Krishna Chaitanya, Ertunc Erdil, Neerav Karani, and Ender Konukoglu.
\newblock Contrastive learning of global and local features for medical image
  segmentation with limited annotations.
\newblock {\em arXiv preprint arXiv:2006.10511}, 2020.

\bibitem{simclr}
Ting Chen, Simon Kornblith, Mohammad Norouzi, and Geoffrey Hinton.
\newblock A simple framework for contrastive learning of visual
  representations.
\newblock In {\em Proc. Int. Conf. Mach. Learn.}, 2020.

\bibitem{mocov2}
Xinlei Chen, Haoqi Fan, Ross Girshick, and Kaiming He.
\newblock Improved baselines with momentum contrastive learning.
\newblock {\em arXiv: Comp. Res. Repository}, 2020.

\bibitem{choy2016universal}
Christopher Choy, JunYoung Gwak, Silvio Savarese, and Manmohan Chandraker.
\newblock Universal correspondence network.
\newblock In {\em Proc. Advances in Neural Inf. Process. Syst.}, pages
  2414--2422, 2016.

\bibitem{cityscapes}
Marius Cordts, Mohamed Omran, Sebastian Ramos, Timo Rehfeld, Markus Enzweiler,
  Rodrigo Benenson, Uwe Franke, Stefan Roth, and Bernt Schiele.
\newblock The cityscapes dataset for semantic urban scene understanding.
\newblock In {\em Proc. IEEE Conf. Comp. Vis. Patt. Recogn.}, pages 3213--3223,
  2016.

\bibitem{imagenet}
Jia Deng, Wei Dong, Richard Socher, Li-Jia Li, Kai Li, and Li Fei-Fei.
\newblock {ImageNet}: A large-scale hierarchical image database.
\newblock In {\em Proc. IEEE Conf. Comp. Vis. Patt. Recogn.}, pages 248--255,
  2009.

\bibitem{context15}
Carl Doersch, Abhinav Gupta, and Alexei Efros.
\newblock Unsupervised visual representation learning by context prediction.
\newblock In {\em Proc. IEEE Int. Conf. Comp. Vis.}, pages 1422--1430, 2015.

\bibitem{dosovitskiy2015flownet}
Alexey Dosovitskiy, Philipp Fischer, Eddy Ilg, Philip Hausser, Caner Hazirbas,
  Vladimir Golkov, Patrick Van Der~Smagt, Daniel Cremers, and Thomas Brox.
\newblock {FlowNet}: Learning optical flow with convolutional networks.
\newblock In {\em Proc. IEEE Int. Conf. Comp. Vis.}, pages 2758--2766, 2015.

\bibitem{voc}
Mark Everingham, Luc Van~Gool, Christopher~KI Williams, John Winn, and Andrew
  Zisserman.
\newblock {The PASCAL Visual Object Classes (VOC) Challenge}.
\newblock {\em Int. J. Comput. Vision}, 88(2):303--338, 2010.

\bibitem{stereoscan}
Andreas Geiger, Julius Ziegler, and Christoph Stiller.
\newblock {StereoScan}: Dense 3d reconstruction in real-time.
\newblock In {\em IEEE Intelligent Vehicles Symp.}, pages 963--968, 2011.

\bibitem{gidaris2018rotations}
Spyros Gidaris, Praveer Singh, and Nikos Komodakis.
\newblock Unsupervised representation learning by predicting image rotations.
\newblock In {\em Proc. Int. Conf. Learn. Representations}, 2018.

\bibitem{gan}
Ian Goodfellow, Jean Pouget-Abadie, Mehdi Mirza, Bing Xu, David Warde-Farley,
  Sherjil Ozair, Aaron Courville, and Yoshua Bengio.
\newblock Generative adversarial nets.
\newblock In {\em Proc. Advances in Neural Inf. Process. Syst.}, pages
  2672--2680, 2014.

\bibitem{GoyalM0M19}
Priya Goyal, Dhruv Mahajan, Abhinav Gupta, and Ishan Misra.
\newblock Scaling and benchmarking self-supervised visual representation
  learning.
\newblock In {\em Proc. IEEE Int. Conf. Comp. Vis.}, 2019.

\bibitem{byol}
Jean-Bastien Grill, Florian Strub, Florent Altch{\'e}, Corentin Tallec, Pierre
  Richemond, Elena Buchatskaya, Carl Doersch, Bernardo~Avila Pires,
  Zhaohan~Daniel Guo, Mohammad~Gheshlaghi Azar, et~al.
\newblock Bootstrap your own latent: A new approach to self-supervised
  learning.
\newblock {\em arXiv: Comp. Res. Repository}, 2020.

\bibitem{halimi2019unsupervised}
Oshri Halimi, Or Litany, Emanuele Rodola, Alex~M Bronstein, and Ron Kimmel.
\newblock Unsupervised learning of dense shape correspondence.
\newblock In {\em Proc. IEEE Conf. Comp. Vis. Patt. Recogn.}, pages 4370--4379,
  2019.

\bibitem{han2020self}
Tengda Han, Weidi Xie, and Andrew Zisserman.
\newblock Self-supervised co-training for video representation learning.
\newblock {\em Proc. Advances in Neural Inf. Process. Syst.}, 33, 2020.

\bibitem{moco}
Kaiming He, Haoqi Fan, Yuxin Wu, Saining Xie, and Ross Girshick.
\newblock Momentum contrast for unsupervised visual representation learning.
\newblock In {\em Proc. IEEE Conf. Comp. Vis. Patt. Recogn.}, 2020.

\bibitem{he2019rethinking}
Kaiming He, Ross Girshick, and Piotr Doll{\'a}r.
\newblock Rethinking imagenet pre-training.
\newblock In {\em Proc. IEEE Int. Conf. Comp. Vis.}, pages 4918--4927, 2019.

\bibitem{he2016deep}
Kaiming He, Xiangyu Zhang, Shaoqing Ren, and Jian Sun.
\newblock Deep residual learning for image recognition.
\newblock In {\em Proc. IEEE Conf. Comp. Vis. Patt. Recogn.}, 2016.

\bibitem{kerl2013dense}
Christian Kerl, J{\"u}rgen Sturm, and Daniel Cremers.
\newblock Dense visual slam for rgb-d cameras.
\newblock In {\em Proc. {IEEE/RSJ} Int. Conf. Intelligent Robots \& Systems},
  pages 2100--2106, 2013.

\bibitem{kim2017fcss}
Seungryong Kim, Dongbo Min, Bumsub Ham, Sangryul Jeon, Stephen Lin, and
  Kwanghoon Sohn.
\newblock {FCSS}: Fully convolutional self-similarity for dense semantic
  correspondence.
\newblock In {\em Proc. IEEE Conf. Comp. Vis. Patt. Recogn.}, pages 6560--6569,
  2017.

\bibitem{kong2016hypernet}
Tao Kong, Anbang Yao, Yurong Chen, and Fuchun Sun.
\newblock {HyperNet}: Towards accurate region proposal generation and joint
  object detection.
\newblock In {\em Proc. IEEE Conf. Comp. Vis. Patt. Recogn.}, pages 845--853,
  2016.

\bibitem{li2019analysis}
Hengduo Li, Bharat Singh, Mahyar Najibi, Zuxuan Wu, and Larry~S. Davis.
\newblock An analysis of pre-training on object detection.
\newblock {\em arXiv: Comp. Res. Repository}, 2019.

\bibitem{li2018detnet}
Zeming Li, Chao Peng, Gang Yu, Xiangyu Zhang, Yangdong Deng, and Jian Sun.
\newblock {DetNet}: Design backbone for object detection.
\newblock In {\em Proc. Eur. Conf. Comp. Vis.}, pages 334--350, 2018.

\bibitem{fpn}
Tsung{-}Yi Lin, Piotr Doll{\'{a}}r, Ross~B. Girshick, Kaiming He, Bharath
  Hariharan, and Serge~J. Belongie.
\newblock Feature pyramid networks for object detection.
\newblock In {\em Proc. IEEE Conf. Comp. Vis. Patt. Recogn.}, 2017.

\bibitem{coco}
Tsung-Yi Lin, Michael Maire, Serge Belongie, James Hays, Pietro Perona, Deva
  Ramanan, Piotr Doll{\'a}r, and Lawrence Zitnick.
\newblock Microsoft {COCO}: Common objects in context.
\newblock In {\em Proc. Eur. Conf. Comp. Vis.}, pages 740--755, 2014.

\bibitem{long2015fully}
Jonathan Long, Evan Shelhamer, and Trevor Darrell.
\newblock Fully convolutional networks for semantic segmentation.
\newblock In {\em Proc. IEEE Conf. Comp. Vis. Patt. Recogn.}, pages 3431--3440,
  2015.

\bibitem{mahajan2018exploring}
Dhruv Mahajan, Ross Girshick, Vignesh Ramanathan, Kaiming He, Manohar Paluri,
  Yixuan Li, Ashwin Bharambe, and Laurens van~der Maaten.
\newblock Exploring the limits of weakly supervised pretraining.
\newblock In {\em Proc. Eur. Conf. Comp. Vis.}, pages 181--196, 2018.

\bibitem{jigsaw}
Mehdi Noroozi and Paolo Favaro.
\newblock Unsupervised learning of visual representations by solving jigsaw
  puzzles.
\newblock In {\em Proc. Eur. Conf. Comp. Vis.}, pages 69--84, 2016.

\bibitem{oord2018representation}
Aaron van~den Oord, Yazhe Li, and Oriol Vinyals.
\newblock Representation learning with contrastive predictive coding.
\newblock {\em arXiv: Comp. Res. Repository}, 2018.

\bibitem{mmseg}
OpenMMLab.
\newblock mmsegmentation.
\newblock \url{https://github.com/open-mmlab/mmsegmentation}, 2020.

\bibitem{inpainting16}
Deepak Pathak, Philipp Krahenbuhl, Jeff Donahue, Trevor Darrell, and Alexei
  Efros.
\newblock Context encoders: Feature learning by inpainting.
\newblock In {\em Proc. IEEE Conf. Comp. Vis. Patt. Recogn.}, pages 2536--2544,
  2016.

\bibitem{pinheiro2020unsupervised}
Pedro Pinheiro, Amjad Almahairi, Ryan~Y Benmaleck, Florian Golemo, and Aaron
  Courville.
\newblock Unsupervised learning of dense visual representations.
\newblock {\em arXiv: Comp. Res. Repository}, 2020.

\bibitem{yolo}
Joseph Redmon, Santosh Divvala, Ross Girshick, and Ali Farhadi.
\newblock {You Only Look Once}: Unified, real-time object detection.
\newblock In {\em Proc. IEEE Conf. Comp. Vis. Patt. Recogn.}, pages 779--788,
  2016.

\bibitem{redmon2017yolo9000}
Joseph Redmon and Ali Farhadi.
\newblock {YOLO9000}: better, faster, stronger.
\newblock In {\em Proc. IEEE Conf. Comp. Vis. Patt. Recogn.}, pages 7263--7271,
  2017.

\bibitem{ren2015faster}
Shaoqing Ren, Kaiming He, Ross Girshick, and Jian Sun.
\newblock {Faster R-CNN}: Towards real-time object detection with region
  proposal networks.
\newblock In {\em Proc. Advances in Neural Inf. Process. Syst.}, pages 91--99,
  2015.

\bibitem{schonberger2016structure}
Johannes~L Schonberger and Jan-Michael Frahm.
\newblock Structure-from-motion revisited.
\newblock In {\em Proc. IEEE Conf. Comp. Vis. Patt. Recogn.}, pages 4104--4113,
  2016.

\bibitem{hrnet}
Ke Sun, Bin Xiao, Dong Liu, and Jingdong Wang.
\newblock Deep high-resolution representation learning for human pose
  estimation.
\newblock In {\em Proc. IEEE Conf. Comp. Vis. Patt. Recogn.}, pages 5693--5703,
  2019.

\bibitem{efficientdet}
Mingxing Tan, Ruoming Pang, and Quoc~V Le.
\newblock {EfficientDet}: Scalable and efficient object detection.
\newblock In {\em Proc. IEEE Conf. Comp. Vis. Patt. Recogn.}, pages
  10781--10790, 2020.

\bibitem{tian2019contrastive}
Yonglong Tian, Dilip Krishnan, and Phillip Isola.
\newblock Contrastive multiview coding.
\newblock {\em arXiv preprint arXiv:1906.05849}, 2019.

\bibitem{tian2020makes}
Yonglong Tian, Chen Sun, Ben Poole, Dilip Krishnan, Cordelia Schmid, and
  Phillip Isola.
\newblock What makes for good views for contrastive learning.
\newblock In {\em NeurIPS}, 2020.

\bibitem{wu2019detectron2}
Yuxin Wu, Alexander Kirillov, Francisco Massa, Wan-Yen Lo, and Ross Girshick.
\newblock Detectron2.
\newblock \url{https://github.com/facebookresearch/detectron2}, 2019.

\bibitem{wu2018unsupervised}
Zhirong Wu, Yuanjun Xiong, Stella Yu, and Dahua Lin.
\newblock Unsupervised feature learning via non-parametric instance
  discrimination.
\newblock In {\em Proc. IEEE Conf. Comp. Vis. Patt. Recogn.}, 2018.

\bibitem{xie2020pointcontrast}
Saining Xie, Jiatao Gu, Demi Guo, Charles~R Qi, Leonidas~J Guibas, and Or
  Litany.
\newblock {PointContrast}: Unsupervised pre-training for 3d point cloud
  understanding.
\newblock {\em arXiv preprint arXiv:2007.10985}, 2020.

\bibitem{zabih1994correspondence}
Ramin Zabih and John Woodfill.
\newblock Non-parametric local transforms for computing visual correspondence.
\newblock In {\em Proc. Eur. Conf. Comp. Vis.}, pages 151--158, 1994.

\bibitem{ZhanX0OL20}
Xiaohang Zhan, Jiahao Xie, Ziwei Liu, Yew{-}Soon Ong, and Chen~Change Loy.
\newblock Online deep clustering for unsupervised representation learning.
\newblock In {\em Proc. IEEE Conf. Comp. Vis. Patt. Recogn.}, 2020.

\bibitem{zhang2016unsupervised}
Chao Zhang, Chunhua Shen, and Tingzhi Shen.
\newblock Unsupervised feature learning for dense correspondences across
  scenes.
\newblock {\em Int. J. Comput. Vision}, 116(1):90--107, 2016.

\bibitem{zhang2016colorful}
Richard Zhang, Phillip Isola, and Alexei Efros.
\newblock Colorful image colorization.
\newblock In {\em Proc. Eur. Conf. Comp. Vis.}, pages 649--666, 2016.

\bibitem{zhao2020makes}
Nanxuan Zhao, Zhirong Wu, Rynson Lau, and Stephen Lin.
\newblock What makes instance discrimination good for transfer learning?
\newblock In {\em Proc. Int. Conf. Learn. Representations}, 2021.

\bibitem{zhou2020cheaperpretrain}
Dongzhan Zhou, Xinchi Zhou, Hongwen Zhang, Shuai Yi, and Wanli Ouyang.
\newblock {Cheaper Pre-training Lunch}: An efficient paradigm for object
  detection.
\newblock {\em arXiv preprint arXiv:2004.12178}, 2020.

\end{thebibliography}
}

\end{document}